%% file: main.tex
\documentclass[journal]{IEEEtran}

\usepackage{graphicx}
\usepackage[utf8]{inputenc}
\usepackage{amsmath}
\usepackage{algorithm}
\usepackage{algpseudocode}
\usepackage{comment}
\usepackage{subcaption}
\usepackage{multirow}
\usepackage{booktabs}
\usepackage{adjustbox}
\usepackage{pgfplots}
\pgfplotsset{compat=1.18}
\usepackage{amssymb,amsfonts}
\usepackage{authblk}
\usepackage[a4paper, total={170mm,257mm}]{geometry}

\usepackage[font=small,labelfont=bf]{caption}

\usepackage{url}
\usepackage{balance}
\usepackage{fancyhdr}
\fancypagestyle{firstpage}
{
    \fancyhead[L]{\footnotesize \textcopyright This work has been submitted to the IEEE for possible publication. Copyright may be
transferred without notice, after which this version may no longer be accessible.}
    \fancyhead[R]{}
}

\begin{document}

\title{CoG-Guided Weight Correction for Fault-Tolerant Deep Neural Networks}

\author[1]{Bahram Parchekani}
\author[1]{Samira Nazari}
\author[1]{Ali Azarpeyvand}
\author[2]{\\Mohammad Hasan Ahmadilivani}
\author[2]{Tara Ghasempouri}
\author[2]{Jaan Raik}

\affil[1]{University of Zanjan, Zanjan, Iran}
\affil[2]{Tallinn University of Technology, Tallinn, Estonia}
\affil[1]{\{bahram.parchekani, samira.nazari, azarpeyvand\}@znu.ac.ir}
\affil[2]{\{mohammad.ahmadilivani, jaan.raik, tara.ghasempouri\}@taltech.ee}
 \vspace{-6mm}

\maketitle
\thispagestyle{firstpage}

\begin{abstract}
Deep Neural Networks (DNNs) used in safety-critical applications are vulnerable to hardware and memory faults that corrupt network weights and degrade reliability. In this paper, we propose a Center of Gravity (CoG) guided weight correction method that restores faulty weights based on their spatial characteristics within each layer. The proposed approach detects and corrects weight faults using distance-aware correction rules, eliminating the need for retraining or architectural modification. 
The effectiveness of the proposed method in terms of the capability of tolerating hardware faults has been evaluated through performing fault injection at different Bit Error Rates (BERs).
Experiments on safety-critical LSTM-based Networks, including StageNet for disease progression tracking and MTFNet for cardiac anomaly detection, demonstrate fault tolerance improvements of up to $230$× and $6.41$×, respectively at a BER of $10^{-3}$, with negligible accuracy loss. When extended to Convolutional Neural Networks (CNNs), the method achieves up to $49.55$× and $20.79$× improvements under comparable fault conditions on ResNet-18 and VGG-16, respectively. To the best of our knowledge, this is the first work to apply the CoG concept to neural network weight tensors for enhancing model reliability. 

\end{abstract}


\IEEEpeerreviewmaketitle

\input{sections/1-introduction}

\input{sections/3-Preliminaries}
\input{sections/4-Methodology}

\input{sections/Results}

\input{sections/6-Conclusion}



\bibliographystyle{IEEEtran}
\bibliography{ref}

\end{document}

%% file: sections/1-introduction.tex
\section{Introduction}


    


    

Machine Learning (ML), particularly Deep Neural Networks (DNNs), have emerged as a powerful solution in a wide range of domains, including image recognition, natural language processing, and autonomous systems \cite{voulodimos2018deep,lin2020mcunet,loni2022tas}. Among these, safety-critical applications such as healthcare and automotive require not only high accuracy but also high dependability \cite{su2023testability,ahmadilivani2024systematic,bolchini2024resilience}. In such domains, hardware reliability is of paramount importance, as faulty predictions may lead to catastrophic consequences.

Hardware reliability is defined as the probability of 
the system producing correct results in the presence of hardware faults. Hardware faults can arise from multiple sources, such as soft errors (induced by high-energy particle strikes), process variations, temperature variations, and aging \cite{ibrahim2020soft,bolchini2022fast}. These faults manifest themselves as bitflips in logic or memory, potentially corrupting intermediate computations or model parameters. Due to technology scaling, fault rates, particularly in memory components, have increased significantly, threatening the reliable deployment of DNN accelerators \cite{ibe2010impact,malekzadeh2021impact,neggaz2019cnns}. 

Within healthcare applications, DNNs are extensively employed for tasks such as disease diagnosis, treatment planning, and anomaly detection \cite{manne2021application,harris2022survey}. These applications often rely on processing time-series data. In this context, Long Short-Term Memory (LSTM) networks are effective due to their ability to retain long-term temporal dependencies through recurrent data processing \cite{lim2021time}. Their success in tasks such as arrhythmia detection, disease stage prediction, and other anomaly monitoring systems makes them highly effective for real-world healthcare applications.

However, the deployment of LSTMs on hardware accelerators exposes them to reliability threats. A single hardware fault may alter a weight or a hidden state within the LSTM cell, leading to incorrect classification or prediction \cite{ahmadilivani2023analysis,nosrati2024analysis}. For example, in disease stage prediction (e.g., cancer) using LSTM models, a single bitflip in memory can misclassify a \textit{critical} stage as \textit{normal}. Such an error prevents timely and relevant treatment, potentially resulting in fatal outcomes. Such scenarios highlight the urgent need for studies on the hardware reliability of LSTMs in healthcare applications, as well as the development of respective fault tolerance mechanisms. As memories are highly susceptible to faults and their content is not overwritten as frequently as buffers and latches in the datapath, protecting DNNs' parameters, which are stored in memory, is necessary to ensure a reliable deployment.



Several studies have investigated the reliability of DNNs in safety-critical applications such as autonomous driving and healthcare \cite{ibrahim2020soft,su2023testability,ahmadilivani2024systematic,bolchini2024resilience}. Despite the extensive use of LSTM networks in these domains, their reliability under hardware faults and the development of cost-effective fault tolerance mechanisms remain relatively underexplored \cite{ahmadilivani2024systematic}.
Recent research has started addressing this gap. 
\cite{ahmadilivani2023analysis} presents a detailed resilience characterization of LSTM parameters, revealing that recurrent weights are the most fault-sensitive components, as they directly influence the memory behavior of LSTMs. The authors further demonstrated that resetting faulty weights to zero can significantly enhance resilience.  

Moreover, \cite{parchekani2024zero} proposes two zero-memory-overhead protection mechanisms for LSTMs: (i) applying range restriction on activation function inputs and (ii) zeroing the most vulnerable weight bits. Their findings indicate that selectively resetting critical bits provides superior protection compared to activation restriction. \cite{vasquez2024soft} applies a Hamming code to the exponent bits of parameters of LSTMs to protect memory against soft errors. 

The existing approaches attempt to detect and correct or mitigate faults, yet their impacts remain limited as they are agnostic to the values of the parameters and their statistical distribution. The spatial patterns in weight matrices can play a critical role in the functionality of a DNN. In this work, by analyzing the spatial patterns of weight matrices, for the first time, we propose a new scheme for correcting erroneous parameters based on the concept of Center of Gravity. We analyze how large-magnitude weights within a layer are located, determine a center point where the values are larger as CoG, and detect and correct the weights corresponding to their distance from CoG.

The contributions of this paper are as follows:

\begin{itemize}
    \item A lightweight detection mechanism is presented in which faulty weights are identified using a golden (fault-free) boundary derived from each layer’s statistical distribution. 
    
    \item A spatial correction strategy is introduced that utilizes the CoG of each weight tensor to guide fault recovery. Faulty weights are corrected according to their spatial distance to the CoG, thereby preserving statistical balance and reducing error propagation.
    
    \item The proposed method is validated on safety-critical LSTM-based networks (StageNet for disease progression tracking and MTFNet for cardiac anomaly detection) using real-world datasets as well as on conventional CNNs (ResNet-18 and VGG-16). 
\end{itemize}

The remainder of the paper is organized as follows. Section \ref{sec:weight-properties} presents the analysis of weight matrix properties and introduces relevant preliminaries, including the definition of the CoG, which provides the foundation for designing effective fault detection and correction strategies. Section \ref{sec:method} introduces the CoG-guided weight correction scheme. Section \ref{sec:experiments} describes the experimental setup, evaluation on LSTM-based and conventional networks, and discusses the results. Finally, Section \ref{sec:conclusion} concludes the paper.

%% file: sections/3-Preliminaries.tex
\section{Statistical Properties of Weight Matrices in DNNs}
\label{sec:weight-properties}






In DNNs, the parameters of each layer are organized as weight matrices (or tensors in convolutional/recurrent layers), whose statistical and spatial characteristics strongly influence their functionality and fault sensitivity. Profiling these features in pre-trained DNNs provides insights into their behavior and guides the design of targeted fault detection and correction strategies. 

In general, the majority of weight values in a layer are distributed within a finite range, such as \([-0.5,+0.5]\) or \([-1,+1]\), depending on the DNN's architecture as well as the initialization and regularization applied during training ~\cite{han2015learning}. Rare outliers with larger magnitudes may still exist, especially in deeper or less regularized DNNs. This bounded distribution reflects the stability of a pre-trained DNN and contributes to numerical stability during inference~\cite{han2015learning}\cite{zhang2016understanding}. The histogram of the weights in a given layer often approximates a normal (Gaussian) distribution, with the majority of weights concentrated near zero. Such a bell-shaped curve results from the optimization process and the application of regularization techniques, such as L2 regularization~\cite{zhang2016understanding}, ensuring stable forward propagation and reducing the risk of exploding or vanishing activations~\cite{D2L-Num-Stab-Initi}\cite{AutoInit-paper}.

    
 Moreover, The average of weights in a layer is typically close to zero, indicating no strong bias in information flow, while their variance remains low, implying tight concentration around the mean ~\cite{glorot2010understanding}\cite{goodfellow2016deep}. 
Such low-variance, zero-mean distributions guarantee the DNNs' reliability during inference~\cite{szegedy2017inception}\cite{he2015delving}. While most weights are small and centered around zero, a smaller subset may have larger magnitudes. These large-magnitude weights often form localized clusters that play a critical role in representing task-specific features. For example, in CNNs, filters with larger weights frequently resemble Gabor-like patterns~\cite{yosinski2015understanding}, which detect edges and textures with specific orientations and spatial frequencies. This combination of global sparsity and locally important weights reflects that 
    only a subset of parameters influences the DNNs' outputs~\cite{frankle1810lottery}\cite{lecun2015deep}.
    




To capture the spatial concentration of large-magnitude values in matrices, a physics-inspired metric, Center of Gravity, is introduced~\cite{cog2023reliability}. In the context of DNNs, CoG can represent the centroid of concentrated weight magnitudes within a layer’s weight tensor, serving as a basis for spatially guided fault detection and correction.

\begin{figure}[h]
    \centering
    \includegraphics[width=0.85\linewidth]{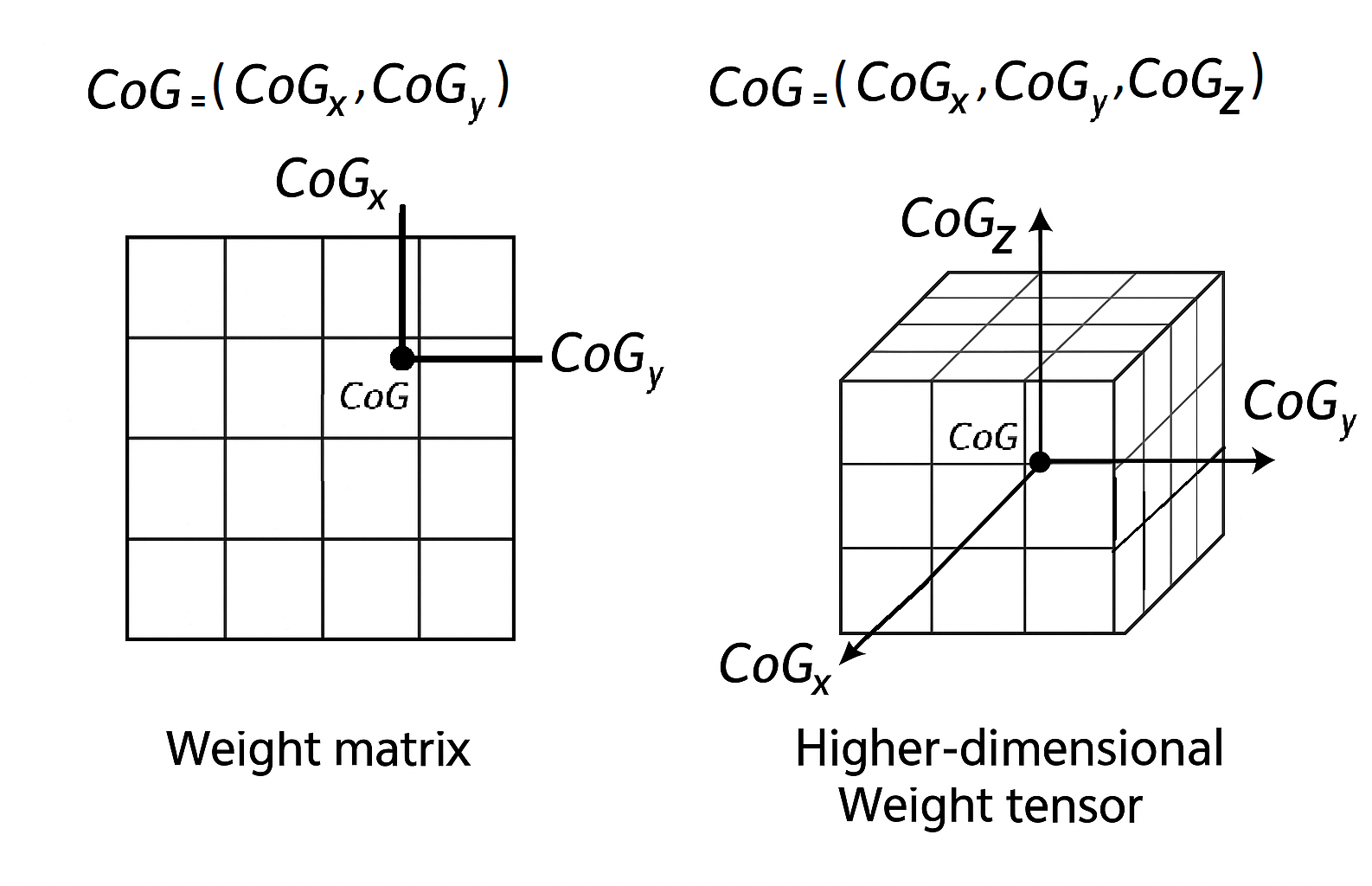}
    \caption{Illustration of the CoG concept for weight matrices. 
    (Left) CoG in a 2D weight matrix. 
    (Right) Extension of the concept to higher-dimensional matrices. 
    }
    \label{fig:cog}
\end{figure}

Mathematically, for a weight matrix \( W \in \mathbb{R}^{m \times n} \) with elements \( w_{ij} \), the row-wise CoG is defined as: 

\begin{equation}
 CoG_x = \frac{\sum_{i=1}^{m} \sum_{j=1}^{n} |w_{ij}| \cdot i}{\sum_{i=1}^{m} \sum_{j=1}^{n} |w_{ij}|},
 \label{eq:cogx}
\end{equation}
where $i$ and $j$ represent the row and column indices, respectively. 
The column-wise CoG is defined as:

\begin{equation}
CoG_y = \frac{\sum_{i=1}^{m} \sum_{j=1}^{n} |w_{ij}| \cdot j}{\sum_{i=1}^{m} \sum_{j=1}^{n} |w_{ij}|}.
\label{eq:cogy}
\end{equation}
The CoG of a 2D matrix corresponds to the point $(CoG_x, CoG_y)$, and the concept naturally extends to higher-rank tensors by computing the weighted average along each dimension (Fig. \ref{fig:cog}).


These definitions assign greater influence to positions with larger-magnitude weights, thereby identifying the central region of the most impactful weights. This technique is analogous to centroid computation in image processing and pattern recognition, where pixel intensities are treated as masses to locate the geometric center of an object \cite{gonzalez2009digital}\cite{han2015learning}.

Prior studies have shown that faults in DNN weights can propagate through layers and significantly degrade inference accuracy, particularly when they affect large-magnitude parameters~\cite{li2017error,reagen2016minerva,zhang2018clipping}. In contrast, small perturbations are often masked by activation nonlinearities and statistical redundancy. Existing fault-tolerance methods rely on statistical profiling of weights but rarely exploit their spatial organization. Since large-magnitude weights tend to cluster in specific regions, identifying these regions enables spatially targeted fault detection and correction. The proposed CoG-based method builds upon this observation to enhance DNN reliability with minimal overhead.











%% file: sections/4-Methodology.tex
\section{Methodology: CoG-based Fault Resilience Enhancement }
\label{sec:method}
\begin{figure}[H]
    \centering
    \includegraphics[width=1\linewidth]{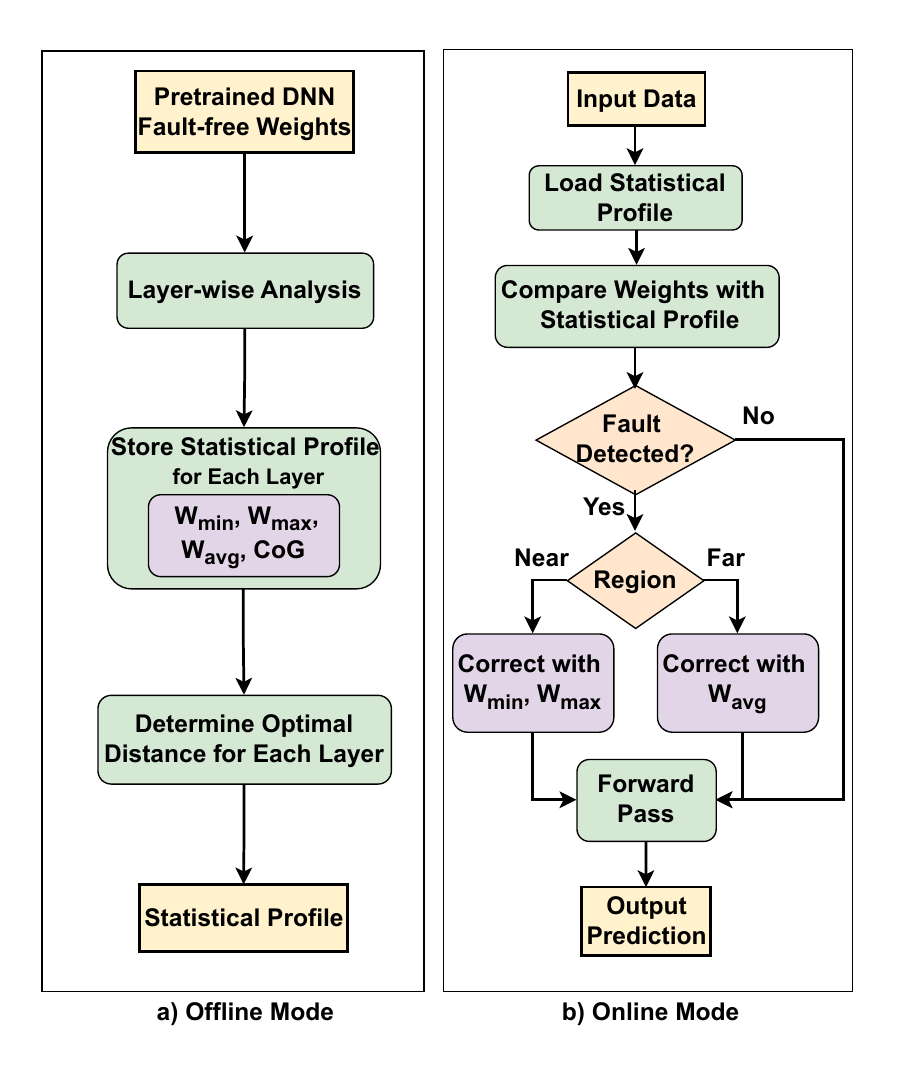}
    \caption{Overview of the CoG-Based Error Detection and Correction Strategy. (a) Offline mode, where a small set of layer-wise reference parameters (minimum, maximum, average, CoG, and optimal distance) is extracted once from the golden pre-trained weights; (b) Online mode, where weight faults are detected and selectively corrected using the precomputed CoG and distance-aware thresholds.}
    \label{fig:cog-overview}
\end{figure}

This section presents the proposed CoG-based method for enhancing the resilience of DNNs against weight faults. Fig. \ref{fig:cog-overview} illustrates the overall workflow, which is divided into two complementary phases: 1) Offline mode: for profiling; 2) Online Mode: for fault detection and correction during inference. 

In the \textbf{Offline Mode}, the DNN’s golden pre-trained weights are analyzed once for each layer to extract a small and fixed set of reference parameters.  Specifically, only five scalar values per layer are stored: the minimum ($w_{min}^l$), maximum ($w_{max}^l$), and average weight values ($w_{avg}^l$), the CoG, and the optimal distance ($d_l^{opt}$) which are explained later in algorithms. These parameters form the golden profile of each layer and serve as the statistical baseline for fault detection, without requiring storage of the full model or weight tensors.

In the \textbf{Online Mode}, the stored statistical profile is used to identify abnormal weights. Corrupted weights that exceed the profiled range are flagged as faulty, while small deviations within the range are tolerated. 
Correction is then based on the CoG, where weights located near the CoG are treated as critical and corrected conservatively using boundary values ($w_{min}^l$ and $w_{max}^l$), while weights farther from the CoG are corrected using the local average of golden weights ($w_{avg}^l$). The corrected weights are subsequently used in the forward pass, ensuring reliable inference without retraining or hardware redundancy.

This dual-phase workflow leverages the pre-profiled statistical and spatial information of each layer to improve reliability during inference. The following subsections present the mathematical formulation of the CoG-based protection scheme and the procedure for determining the optimal correction distance.

\subsection{Fault Detection in Weights}


Fault detection relies on the statistical profile established during Offline Mode, where only a small set of layer-wise reference parameters is stored.
Each weight is compared with the range $[w^l_{min},w^l_{max}]$; those exceeding it are flagged as faulty:

\begin{equation}
E_{ij} =
\begin{cases}
1, & (w_{ij} < w^l_{min}) \text{ or } (w_{ij} > w^l_{max}) \\
0, & \text{otherwise}
\end{cases}
\label{eq:E}
\end{equation}

Here, $E_{ij} = 1$ indicates a likely memory fault, while $E_{ij} = 0$ denotes a valid weight.
This simple element-wise test efficiently detects faults without retraining or hardware modification.


\subsection{CoG-Based Protection}

\begin{figure}[b]
    \centering
    \includegraphics[width=0.85\linewidth]{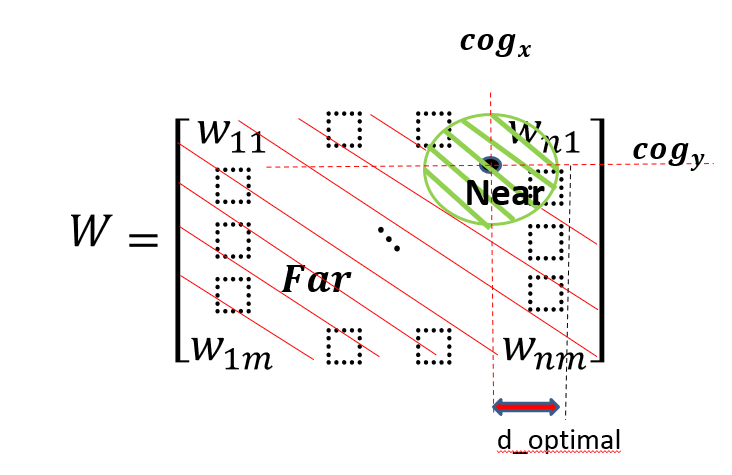}
    \caption{The weight matrix is divided into two regions, Near and Far from the CoG.}
    \label{fig:protectioncog}
\end{figure}

Once faulty weights are detected, correction is guided by their spatial proximity to the CoG of each layer. The CoG and statistical profile are obtained from the Offline profiling phase. As shown in Fig.~\ref{fig:protectioncog}, the weight matrix is divided into Near and Far regions relative to the CoG, based on a tunable distance, $d_{opt}$ (the green circle).

Weights closer to the CoG typically have larger magnitudes and impose a more significant influence on DNN outputs. Hence, the correction strategy is applied selectively: preserving high-impact regions near the CoG while maintaining statistical balance through the far values.

\subsubsection{Region-Specific Correction Rules}
If a weight is detected as faulty, the correction method depends on the region it belongs to:

\textbf{a. Near-CoG Region:}
If a faulty weight \(w'_{ij}\) is located in the region near the CoG, it is considered critical due to its proximity to large-magnitude weights. It is formally defined by Eq. \eqref{eq:near}, which ensures that weights in the region \textit{Near} are corrected conservatively by clamping them to the valid boundaries.

\begin{equation} 
w_{ij} = 
\begin{cases} 
w^l_{min}  & \text{if } w'_{ij} < w^l_{min} \\
w^l_{max} & \text{if } w'_{ij} > w^l_{max} 
\end{cases} 
\label{eq:near}
\end{equation}

\textbf{b. Far-from-CoG Region:}
The faulty weights that fall in the region far from the CoG, which are typically small in magnitude and less critical, are replaced by the profiled average value of all weights in that layer, as previously obtained in offline mode.



\subsubsection{Optimization of the Distance} \label{sec:optimization}



The optimal distance $d_{opt}$ determines the boundary between Near and Far regions.
It is tuned to maximize post-correction accuracy through iterative fault injection and evaluation. Two search strategies are explored: exhaustive and binary.

\textbf{a. Exhaustive Search:}
To determine the optimal distance from CoG, $d_l^{opt}$ for each layer $l$, a layer-wise exhaustive search is performed as summarized in Algorithm~\ref{alg:iterative_search}. For each layer, the statistical profile is extracted, including minimum, maximum, average weight values, and the CoG coordinates (lines 2–3). The maximum search distance $d^l_{max}$ is computed based on the Euclidean distance of all weights from the CoG. Each candidate distance from $1$ to $d^l_{max}$ is then evaluated by performing $N$ independent fault injection experiments (lines 4–12). 

In each iteration, faults are injected into weights with a constant BER, the CoG-based protection is applied, and the model is evaluated on the test data. The metric for each trial is stored and averaged over $N$ iterations to obtain the average performance for that distance (line 13). After evaluating all candidate distances, the distance yielding the highest average metric is selected as the optimal $d_l^{opt}$ (line 15). The process is repeated for all layers.


\begin{algorithm}
\caption{Exhaustive Search}
\label{alg:iterative_search}
\begin{algorithmic}[1]
\Require Test data, pre-trained DNN with $L$ layers, $N$ iterations, $BER$
\Ensure Optimal thresholds $\{d_1^{opt}, d_2^{opt}, \dots, d_L^{opt}\}$
\For{$l = 1$ to $L$}
    \State $StatisticalProfile_l \gets [W^l_{min}, W^l_{max}, W^l_{avg}, CoG^l]$
    \State $d^l_{max} \gets MaxDistance(GoldenWeight^l , CoG^l)$
    \For{$d = 0$ to $d^l_{max}$}
        \State Initialize \textit{metric\_list}
        \For{$n = 1$ to $N$}
            \State Fault\_Injection($GoldenWeight^l$, $BER$)
            \State Protection\_CoG($StatisticalProfile^l$, $d$)
            \State \textit{metric} $\gets$ Evaluate(DNN, Test data, d)
            \State Append \textit{metric} to \textit{metric\_list}
            \State $Weight_l \gets GoldenWeight^l$
        \EndFor
        \State Compute average metric for $d$ and layer $l$
    \EndFor
    \State $d^l_{opt} \gets \arg\max(\text{average metrics})$
\EndFor
\end{algorithmic}
\end{algorithm}

\textbf{b. Binary Search:} 
To reduce the computational cost of exhaustive search, a binary search approach is used to determine $d_l^{opt}$ through Algorithm~\ref{alg:binary_search}. The statistical profile is extracted for each layer (line 2), and the search bounds $d_{min}^l$ and $d_{max}^l$ are initialized (lines 3 and 4). The model performance is first evaluated at the boundaries (lines 5 and 6). Then, the algorithm iteratively computes the midpoint distance, applies protection, evaluates the model, and updates the search bounds depending on which half shows better performance (lines 7–13). This loop continues until the distance interval is smaller than one distance unit (i.e., a single discrete distance step in the binary search), or the performance improvement is below a specified threshold $\theta$. Finally, the endpoint with higher performance is selected as the optimal $d_l^{opt}$ (line 16). This procedure ensures similar resilience to exhaustive search while significantly reducing the number of evaluations.

\begin{algorithm}
\caption{Binary Search for Optimal Distance}
\label{alg:binary_search}
\begin{algorithmic}[1]
\Require Test data, pre-trained DNN with $L$ layers, $N$ iterations, $BER$, threshold $\theta$
\Ensure Optimal distances $\{d_1^{opt}, d_2^{opt}, \dots, d_L^{opt}\}$
\For{$l = 1$ to $L$}
    \State $StatisticalProfile_l \gets [W^l_{min}, W^l_{max}, W^l_{avg}, CoG^l]$
    \State $d^l_{max} \gets MaxDistance(GoldenWeight^l, CoG^l)$
    \State $d^l_{min} \gets 0$
    \State $metric(d_{min}^l)$ $\gets$ Evaluate(DNN, Test data, $d_{min}^l$)
    \State $metric(d_{max}^l)$ $\gets$ Evaluate(DNN, Test data, $d_{max}^l$)
    \While{$(d_{max}^l - d_{min}^l \ge 1)$ \textbf{and} $(|metric(d_{min}^l) - metric(d_{max}^l)| \ge \theta)$}
        \State $d_{mid}^l \gets \lfloor (d_{min}^l + d_{max}^l) / 2 \rfloor$
        \State $metric(d_{mid}^l)$ $\gets$ Evaluate(DNN, Test data, $d_{mid}^l$)
        \If{$(metric(d_{mid}^l) > metric(d_{min}^l))$ \textbf{and} $(metric(d_{mid}^l) > metric(d_{max}^l))$}
            \State $d_{max}^l \gets d_{mid}^l$
        \Else
            \State $d_{min}^l \gets d_{mid}^l$
        \EndIf
    \EndWhile
    \State $d_l^{opt} \gets \arg\max \{ metric(d_{min}), metric(d_{max}) \}$
\EndFor
\end{algorithmic}
\end{algorithm}

%% file: sections/Results.tex
\section{Experimental Results}
\label{sec:experiments}

\subsection{Experimental Setup}
We evaluate the proposed method on four deep neural networks (Table~\ref{tab:model_summary}). \textbf{StageNet} is a stage-aware binary classification model for health risk prediction on the MIMIC-III dataset~\cite{mimiciii}, using LSTM layers to capture temporal dependencies. \textbf{MTFNet} is a multi-label ECG classification model with BiLSTM layers, pretrained on datasets CPSC, CPSC\_Extra~\cite{cpsc2018}, and Shaoxing~\cite{shaoxing2020}, predicting 26 cardiac conditions. To demonstrate the generalizability of the proposed method beyond temporal data and LSTMs, we include two standard CNNs for image classification: \textbf{VGG-16} and \textbf{ResNet-18}, trained on CIFAR-10.  

\begin{table*}[htbp]
\centering
\footnotesize
\caption{Summary of DNN architectures, number of parameters, and baseline (fault-free) performance.}
\label{tab:model_summary}
\begin{tabular}{|l|c|c|c|c|}
\hline
\textbf{Model} & \textbf{Layers} & \textbf{Total Weights} & \textbf{Evaluation Metrics} & \textbf{Baseline Values} \\
\hline
MTFNet     & LSTM + CONV + FC & 477,120 & AUROC / AUPRC / Challenge Metric & 97.90 / 86.20 / 88.50 \\
StageNet   & LSTM + CONV + FC & 2,262,714 & AUROC / Accuracy & 79.21 / 94.94 \\
ResNet-18  & CONV + FC & 11,242,368 & AUROC / AUPRC / Accuracy & 99.47 / 96.69 / 90.80 \\
VGG-16     & CONV + FC & 15,244,096 & AUROC / AUPRC / Accuracy & 99.41 / 96.63 / 91.43 \\
\hline
\end{tabular}
\end{table*}

Fault resilience is evaluated by randomly injecting bitflips into the 32-bit floating-point weights at different BERs. Each experiment is repeated 1000 times to ensure statistically reliable results, and the average performance is reported. Fault injection is conducted in two types of experiments in this work:  
\begin{itemize}
    \item Layer-wise Distance Optimization: Faults are injected into a target layer to determine the optimal correction distance $d^{l}_{opt}$ using the proposed algorithms.
    \item Full-network Fault Resilience: Faults are injected across all weights of the DNN to evaluate overall reliability.
\end{itemize}

Performance is measured using four metrics: AUROC, which quantifies class separability by comparing true and false positive rates; AUPRC, which highlights precision–recall trade-offs in imbalanced datasets; Accuracy, which reflects the overall prediction correctness; and Challenge Metric, which determines clinically reasonable predictions in multi-label cardiac diagnosis . Since AUROC and AUPRC are common across all evaluated models, the unified reliability metric is denoted as AAA (Average AUROC–AUPRC or Accuracy). All metrics are normalized relative to the fault-free baseline and averaged to form a unified reliability metric (AAA). Experiments are implemented in PyTorch and executed on a system with a GeForce RTX 4070 GPU.


\subsection{Optimized Distance Evaluation}
\begin{figure*}[bhtp]
    \centering
    \begin{subfigure}[b]{0.32\textwidth}
        \includegraphics[width=\textwidth]{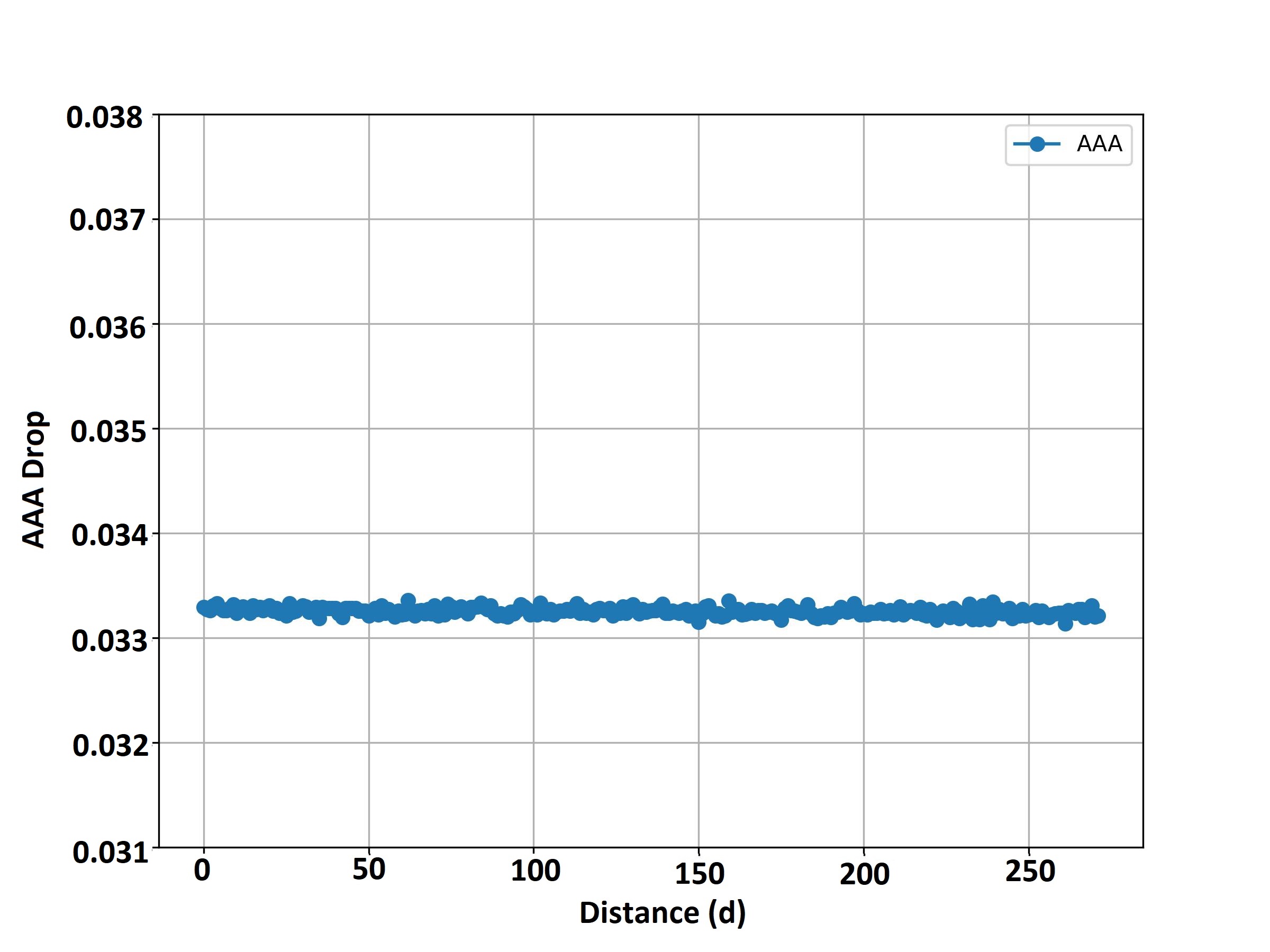}
        \caption{BER=$10^{-4}$}
        \label{fig:sub3}
    \end{subfigure}
    \begin{subfigure}[b]{0.32\textwidth}
        \includegraphics[width=\textwidth]{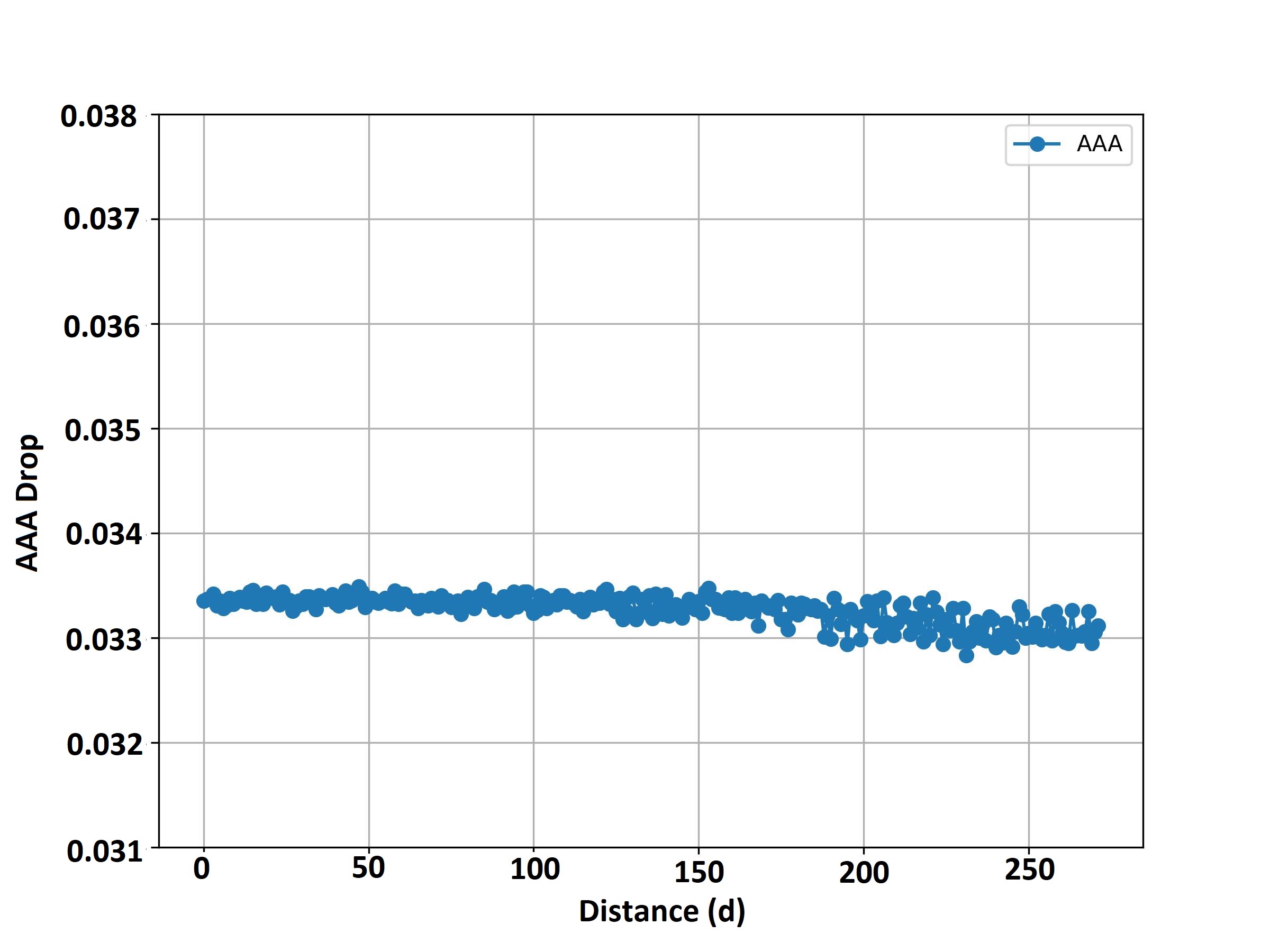}
        \caption{BER=$10^{-3}$}
        \label{fig:sub4}
    \end{subfigure}
    \begin{subfigure}[b]{0.32\textwidth}
        \includegraphics[width=\textwidth]{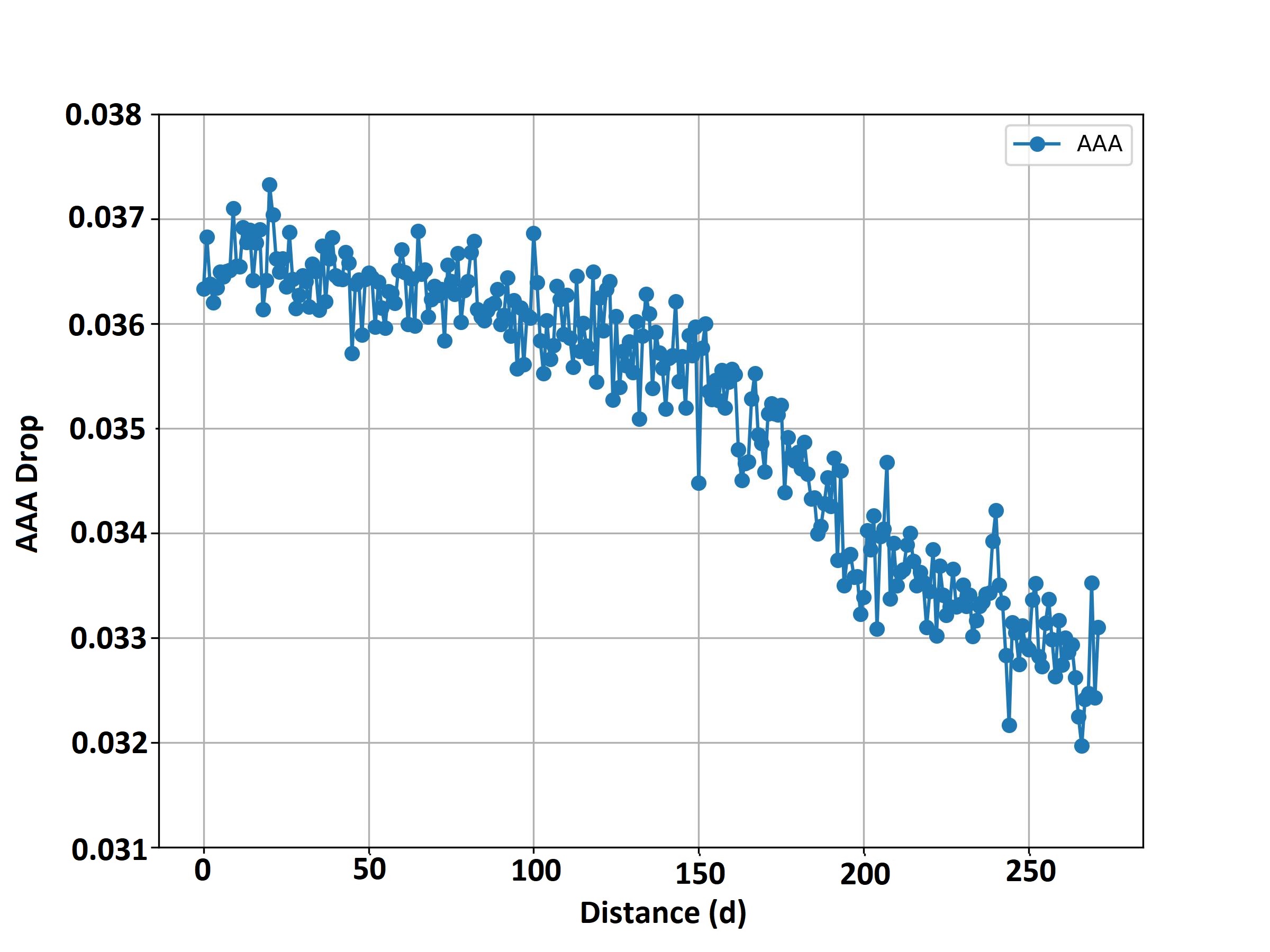}
        \caption{BER=$10^{-2}$}
        \label{fig:sub5}
    \end{subfigure}

    \caption{Variation of AAA drop in the cgrk layer with distance from the CoG under different BERs.}
    \label{fig:cgrk-bers}
\end{figure*}

To specify an effective BER in Algorithm~\ref{alg:iterative_search}, we explore different BERs and examine the performance metric. 
Figure~\ref{fig:cgrk-bers} illustrates the variation of the average performance degradation (AAA drop) in the \textit{cgrk} layer in Stagenet as a representative, with a range of distances from the CoG under different BERs. The curves demonstrate that at the largest BER, the variation of the AAA drop is larger than at lower BERs. 
This observation confirms that using the highest BER for distance optimization can guarantee identifying the optimal distance for protection thresholding and leads to more resilient DNNs. In the distance optimization experiments, we consider BER=$10^{-2}$.

\begin{table}[t]
\footnotesize
\centering
\caption{Layer-wise optimal distances, layer sizes, and maximum possible distances for StageNet using exhaustive search.}
\label{tab:stagenet_exhaustive}
\begin{tabular}{lccc}
\toprule
\textbf{Layer}  & \textbf{Size} & \textbf{Max Distance} & \textbf{Optimal Distance}\\
\midrule
fmrk  & [3, 385]    & 198 & 190 \\
imrk  & [3, 385]    & 195 & 29\\
fgrk  & [384, 385]  & 281 & 277\\
igrk  & [384, 385]  & 279 & 245\\
ogrk  & [384, 385]  & 275 & 253\\
cgrk  & [384, 385]  & 272 & 267\\
fmk     & [3, 94]     & 50 & 1 \\
imk    & [3, 94]     & 50  & 11\\
fgk   & [384, 94]   & 207 & 180\\
igk    & [384, 94]   & 202 & 22\\
ogk     & [384, 94]   & 199 & 5\\
cgk    & [384, 94]   & 201 & 12\\
cnn   & [384, 384, 10] & 283 & 275\\
\bottomrule
\end{tabular}
\end{table}

\begin{table}[htbp]
\centering
\caption{Layer-wise optimal distances, layer sizes, and maximum possible distances for MTFNet using exhaustive search.}
\label{tab:mtfnet_exhaustive}
\begin{tabular}{lccc}
\toprule
\textbf{Layer}  & \textbf{Size} & \textbf{Max Distance} & \textbf{Optimal Distance}\\
\midrule
Conv\_0     & [64, 64, 3]   & 44  & 3\\
Conv\_1    & [64, 64, 3]   & 31  & 22\\
BN          & [64]          & 44  & 4\\
ih          & [256, 128]    & 144 & 5\\
hh         & [256, 64]     & 132 & 10\\
ih\_rev    & [256, 128]    & 144 & 18\\
hh\_rev    & [256, 64]     & 132 & 10\\
Norm        & [128]         & 65  & 0\\
\bottomrule
\end{tabular}
\end{table}

The optimal protection distances for each layer of StageNet and MTFNet are determined using the exhaustive search (Algorithm~\ref{alg:iterative_search}) at the worst-case fault–injection scenario, where severe weight perturbations lead the model to produce highly incorrect outputs, ensuring that the obtained thresholds remain effective under all smaller BERs. The detailed layer-wise optimal distances obtained from exhaustive search are presented in Table~\ref{tab:stagenet_exhaustive} and Table~\ref{tab:mtfnet_exhaustive} for StageNet and MTFNet, respectively.  

Layer names are abbreviated as follows: f, i, o, and c denote forget, input, output, and update gates, respectively; r indicates recurrent weights and k kernel weights (e.g., fmrk: forget-gate recurrent kernel). Conv refers to convolutional layers, BN to batch normalization, and Norm to normalization layers. ih and hh correspond to input-to-hidden and hidden-to-hidden weights of the forward BiLSTM, while the suffix rev denotes the backward direction.

Both architectures comprise pipelines containing Linear, Convolutional, and LSTM layers. During analysis, the Linear layers did not exhibit any consistent or meaningful trend in their optimal protection distances; therefore, the exhaustive search was applied only to the LSTM and Convolutional layers. For the Linear layers, the distance value is set to zero, indicating that in the presence of a weight error, the correction falls back to using the mean value of that layer’s weights.

To reduce the computational cost of exhaustive search, Algorithm~\ref{alg:binary_search} introduces a binary search strategy that efficiently approximates the optimal distances.  
Table~\ref{tab:threshold-error} presents the deviation of binary search results from the exhaustive search, while Table~\ref{tab:alg-time} summarizes the corresponding speedup.

\begin{table}[b]
\centering
\caption{Error (\%) for different thresholds in Algorithm~2.}
\label{tab:threshold-error}
\begin{tabular}{c c c}
\hline
\textbf{Network} & \textbf{Threshold ($th$)} & \textbf{Error (\%)} \\
\hline
\multirow{2}{*}{StageNet} & 0.1  & 24.06 \\
 & 0.01 & 9.70  \\
 \hline
\multirow{2}{*}{MTFNet}   & 0.1  & 7.40  \\
   & 0.01 & 5.04  \\
\hline
\end{tabular}
\end{table}

\begin{table}[htbp]
\centering
\caption{Comparison of computation time (in hours) between Algorithm~1 and Algorithm~2 with $th=0.01$ for obtaining optimal distance, along with the achieved speedup.}
\label{tab:alg-time}
\begin{tabular}{|c|c|c|c|}
\hline
\textbf{Network} & \textbf{Algorithm} & \textbf{Time (h)} & \textbf{Speedup} \\
\hline
\multirow{2}{*}{StageNet} & 1 & 29.75 & \multirow{2}{*}{4.46$\times$} \\
                          & 2 & 6.67  &                       \\
\hline
\multirow{2}{*}{MTFNet}   & 1 & 76.30 & \multirow{2}{*}{5.07$\times$} \\
                          & 2 & 15.02 &                      \\
\hline
\end{tabular}
\end{table}
For StageNet, the binary search achieves only a $9.70\%$ deviation from the exhaustive method at $th=0.01$, and for MTFNet, the deviation decreases to $5.04\%$.  
Despite these minor deviations, Algorithm~2 reduces the overall search time by more than $4.46\times$ for StageNet and $5.07\times$ for MTFNet, demonstrating that the binary approach effectively balances accuracy and efficiency.  

Figures~\ref{fig:stgage_alg1,2} and~\ref{fig:mtf_alg1,2} provide a layer-wise comparison of the optimal distances obtained by Algorithm~1 (exhaustive reference) and Algorithm~2 with $th=0.1$ and $th=0.01$ for StageNet and MTFNet, respectively.  
The results show that the smaller threshold ($th=0.01$) produces optimal distances that more closely match the exhaustive search outcomes, further validating the accuracy of the binary search approximation.




\begin{figure}[b]
    \centering
    \includegraphics[width=1\linewidth]{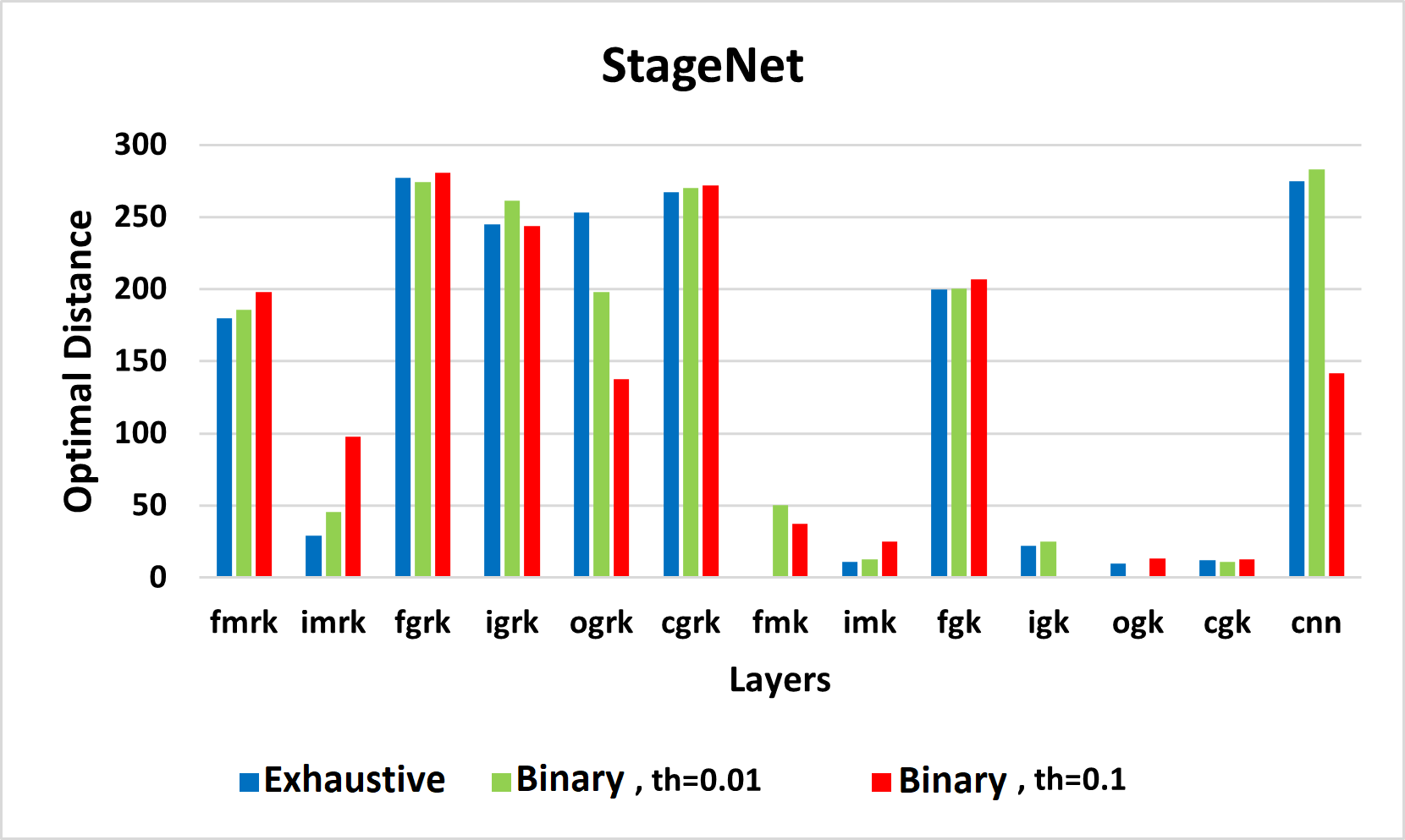}
    \caption{Layer-wise comparison of optimal distance obtained by Algorithm~1 (reference) and Algorithm~2 with $th=0.1$ and $th=0.01$ for StageNet.}
    \label{fig:stgage_alg1,2}
\end{figure}

\begin{figure}[htbp]
    \centering
    \includegraphics[width=1\linewidth]{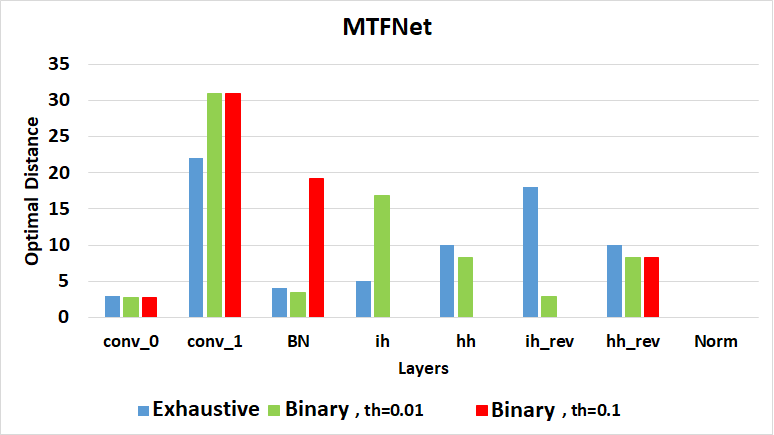}
    \caption{Layer-wise comparison of optimal distance obtained by Algorithm~1 (reference) and Algorithm~2 with $th=0.1$ and $th=0.01$ for MTFNet.}
    \label{fig:mtf_alg1,2}
\end{figure}


\subsection{Fault Resilience Evaluation of the CoG-based protection}
\begin{table*}[htbp]
\centering
\caption{Comparison of AAA drop under different weight reliability protections. Lower values indicate better reliability against fault injection.}
\label{tab:dropaaa-combined}

\begin{subtable}{0.48\textwidth}
\centering
\caption{StageNet (MIMIC-III)}
\begin{tabular}{lrrrrr}
\hline
Protection & 1E-07 & 1E-06 & 1E-05 & 1E-04 & 1E-03 \\
\hline
$CoG_e$ & -2.38 & -2.34 & \textbf{-2.32} & -2.06 & -1.89 \\
$CoG_b$ & \textbf{-2.39} & -2.34 & \textbf{-2.32} & -2.32 & \textbf{-2.23} \\
Average \cite{ahmadilivani2023analysis}& -2.22 & -2.22 & -2.20 & -2.06 & -1.68 \\
MinMax \cite{chen2021low} & -2.25 & -2.27 & -2.31 & -2.29 & -1.69 \\
WBC \cite{parchekani2024zero} & \textbf{-2.39} & \textbf{-2.36} & \textbf{-2.32} & \textbf{-2.36} & -2.04 \\
AVC \cite{parchekani2024zero} & 2.16 & 16.18 & 16.39 & 9.21 & 27.09 \\
NPrt & 1.66 & 16.99 & 16.41 & 14.52 & 32.16 \\
\hline
\end{tabular}
\end{subtable}
\hfill
\begin{subtable}{0.48\textwidth}
\centering
\caption{MTFNet (CPSC)}
\begin{tabular}{lrrrrr}
\hline
Protection & 1E-07 & 1E-06 & 1E-05 & 1E-04 & 1E-03 \\
\hline
$CoG_e$ & \textbf{-0.07} & \textbf{-0.05} & \textbf{-0.07} & \textbf{0.50} & \textbf{7.85} \\
$CoG_b$ & -0.04 & -0.02 & -0.02 & 0.60 & 7.98 \\
Average \cite{ahmadilivani2023analysis}& -0.05 & -0.03 & -0.01 & 1.22 & 9.86 \\
MinMax \cite{chen2021low} & -0.02 & -0.03 & -0.01 & 0.99 & 12.85 \\
WBC \cite{parchekani2024zero} & 0.03 & 0.02 & 1.26 & 15.24 & 59.12 \\
AVC \cite{parchekani2024zero} & 0.03 & 0.32 & 3.59 & 32.88 & 69.62 \\
NPrt & 0.69 & 7.68 & 52.10 & 78.06 & 78.06 \\
\hline
\end{tabular}
\end{subtable}

\vspace{0.5em}

\begin{subtable}{0.48\textwidth}
\centering
\caption{MTFNet (CPSCExtra)}
\begin{tabular}{lrrrrr}
\hline
Protection & 1E-07 & 1E-06 & 1E-05 & 1E-04 & 1E-03 \\
\hline
$CoG_e$ & 0.02 & \textbf{0.11} & 0.80 & 2.29 & 11.20 \\
$CoG_b$ & \textbf{-0.02} & 0.24 & \textbf{0.35} & \textbf{2.15} & \textbf{11.12} \\
Average \cite{ahmadilivani2023analysis} & 1.26 & 1.32 & 1.46 & 2.50 & 11.36 \\
MinMax \cite{chen2021low} & -0.01 & 0.12 & 0.76 & 3.09 & 15.76 \\
WBC \cite{parchekani2024zero} & 0.09 & 0.30 & 2.78 & 16.39 & 46.37 \\
AVC \cite{parchekani2024zero} & 0.06 & 0.85 & 5.91 & 29.88 & 55.59 \\
NPrt & 0.60 & 5.80 & 42.74 & 63.14 & 63.14 \\
\hline
\end{tabular}
\end{subtable}
\hfill
\begin{subtable}{0.48\textwidth}
\centering
\caption{MTFNet (Shaoxing)}
\begin{tabular}{lrrrrr}
\hline
Protection & 1E-07 & 1E-06 & 1E-05 & 1E-04 & 1E-03 \\
\hline
$CoG_e$ & 0.04 & 0.04 & \textbf{0.11} & \textbf{0.54} & \textbf{10.15} \\
$CoG_b$ & 0.05 & 0.05 & 0.13 & 0.99 & 10.36 \\
Average \cite{ahmadilivani2023analysis} & 0.09 & 0.07 & 0.12 & 1.07 & 10.93 \\
MinMax \cite{chen2021low} & \textbf{0.00} & \textbf{0.02} & \textbf{0.11} & 1.63 & 17.94 \\
WBC \cite{parchekani2024zero} & 0.13 & 0.13 & 1.91 & 20.71 & 65.66 \\
AVC \cite{parchekani2024zero} & 0.04 & 0.49 & 4.84 & 39.12 & 75.69 \\
NPrt & 0.81 & 7.75 & 59.45 & 89.39 & 89.39 \\
\hline
\end{tabular}
\end{subtable}

\end{table*}
Using the optimal distances derived above, two CoG-based protection schemes, $CoG_e$ (exhaustive) and $CoG_b$ (binary search), are evaluated against several baseline and statistical approaches: Average~\cite{ahmadilivani2023analysis}, MinMax~\cite{chen2021low}, Weight Bit Clipping (WBC)~\cite{parchekani2024zero}, and Activation Value Clipping (AVC)~\cite{parchekani2024zero}.   
The unprotected model (NPrt) serves as the baseline reference.  

Specifically, the Average method replaces detected faulty weights with the mean value of their respective layer, while MinMax substitutes them with the marginal minimum or maximum values within the same layer.  
WBC constrains weight magnitudes to a predefined safe range to mitigate the effect of bit flips, and AVC applies similar clipping to activation values to suppress extreme outputs caused by faults.  
In contrast, the proposed CoG-based methods leverage the centroid of the weight distribution to characterize normal behavior and perform fine-grained, adaptive correction based on spatial deviation from this centroid.

Table~\ref{tab:dropaaa-combined} illustrates the degradation in the average performance metric (AAA) under the no-protection (NPrt) scenario for all models across varying BERs.  
Table~\ref{tab:dropaaa-combined} compares the average performance degradation (AAA drop) across StageNet and MTFNet for all protection schemes.  
A lower AAA drop indicates stronger fault resilience and better preservation of functional accuracy.  
As shown, both $CoG_e$ and $CoG_b$ maintain nearly zero AAA drop across all tested BERs, confirming their robustness under severe weight perturbations.  
Additionally, Table~\ref{tab:dropaaa-combined} highlights the marginal performance difference between $CoG_e$ and $CoG_b$, demonstrating that the binary variant achieves nearly identical reliability with substantially reduced computational cost. Across all evaluated DNNs, CoG consistently improves the AAA-drop metric compared to the unprotected model (NPrt), with gains ranging from \textbf{34\%} in StageNet to \textbf{70\%}, \textbf{52\%}, and \textbf{79\%} across the three MTFNet datasets, respectively, corresponding to an overall improvement range of \textbf{34\%--79\%} throughout the networks.

\subsection{Comparison with State-of-the-Art Methods}

Table~\ref{tab:fault-mitigation} compares the effectiveness of different protection schemes in mitigating critical faults, quantified as the reduction ratio of SDC\textsubscript{critical} and DUE relative to the unprotected baseline.  
CoG-based protection consistently achieves the highest mitigation across all models: it provides a 230.4$\times$ reduction in StageNet, 6.41$\times$ in MTFNet, 20.79$\times$ in VGG16, and 49.55$\times$ in ResNet18.
Methods such as Average and MinMax provide limited fault masking but lack awareness of spatial distribution, leading to over-smoothing of sensitive weights.  

\begin{table}[b]
\centering
\caption{Error mitigation ratio of protection methods at BER=$10^{-3}$, showing how much each scheme reduces SDC$_{critical}$+DUE compared to the unprotected baseline. larger values indicate stronger fault mitigation.}
\label{tab:fault-mitigation}
\begin{tabular}{l c c c c c}
\hline
\textbf{Model} & \textbf{CoG-based} & \textbf{\textbf{\begin{tabular}[c]{@{}c@{}}Average\\ \cite{ahmadilivani2023analysis}\end{tabular}}} & \textbf{\textbf{\begin{tabular}[c]{@{}c@{}}MinMax\\ \cite{chen2021low}\end{tabular}}} & \textbf{\textbf{\begin{tabular}[c]{@{}c@{}}WBC\\ \cite{parchekani2024zero}\end{tabular}}} & \textbf{\textbf{\begin{tabular}[c]{@{}c@{}}AVC\\ \cite{parchekani2024zero}\end{tabular}}} \\ \hline
StageNet & 230.44 & 167.52 & 127.98 & 153.58 & 1.49 \\
MTFNet & 6.41 & 4.93 & 3.41 & 1.07 & 1.01 \\
VGG16 & 20.79 & 19.65 & 13.13 & 6.24 & 1.00 \\
ResNet18 & 49.55 & 42.67 & 28.90 & 11.57 & 1.00 \\
\hline
\end{tabular}
\end{table}
WBC and AVC improve reliability under mild fault rates but experience significant degradation as the BER increases.  

In contrast, CoG-based protection leverages spatial information through the CoG and distance-adaptive thresholds to selectively correct faulty weights according to their geometric relevance, preserving both accuracy and structural consistency.  
Although the proposed CoG-based schemes incur larger runtime overhead due to layer-wise CoG computation and distance-based correction, they achieve substantially better accuracy under the same fault conditions.  
This trade-off is quantitatively summarized in Table~\ref{tab:runtime-overhead}, which reports the runtime overhead of different protection methods for single-bit error correction.  
\begin{table}[htbp]
\centering
\caption{Runtime overhead of protection methods (single-bit error correction).}
\label{tab:runtime-overhead}
\begin{tabular}{l c c}
\hline
\textbf{Method} & \textbf{Correction operations} & \textbf{Latency ($\mu s$)} \\ \hline
CoG-based & 3 & 182 \\
Average \cite{ahmadilivani2023analysis} & 1 & 44 \\
MinMax\cite{chen2021low} & 2 & 55 \\
WBC \cite{parchekani2024zero} & 0 & 0 \\
AVC \cite{parchekani2024zero} & 2 & 55 \\
\hline
\end{tabular}
\end{table}

\begin{figure*}[htbp]
    \centering
    \subfloat[Golden model]{\includegraphics[width=0.32\textwidth]{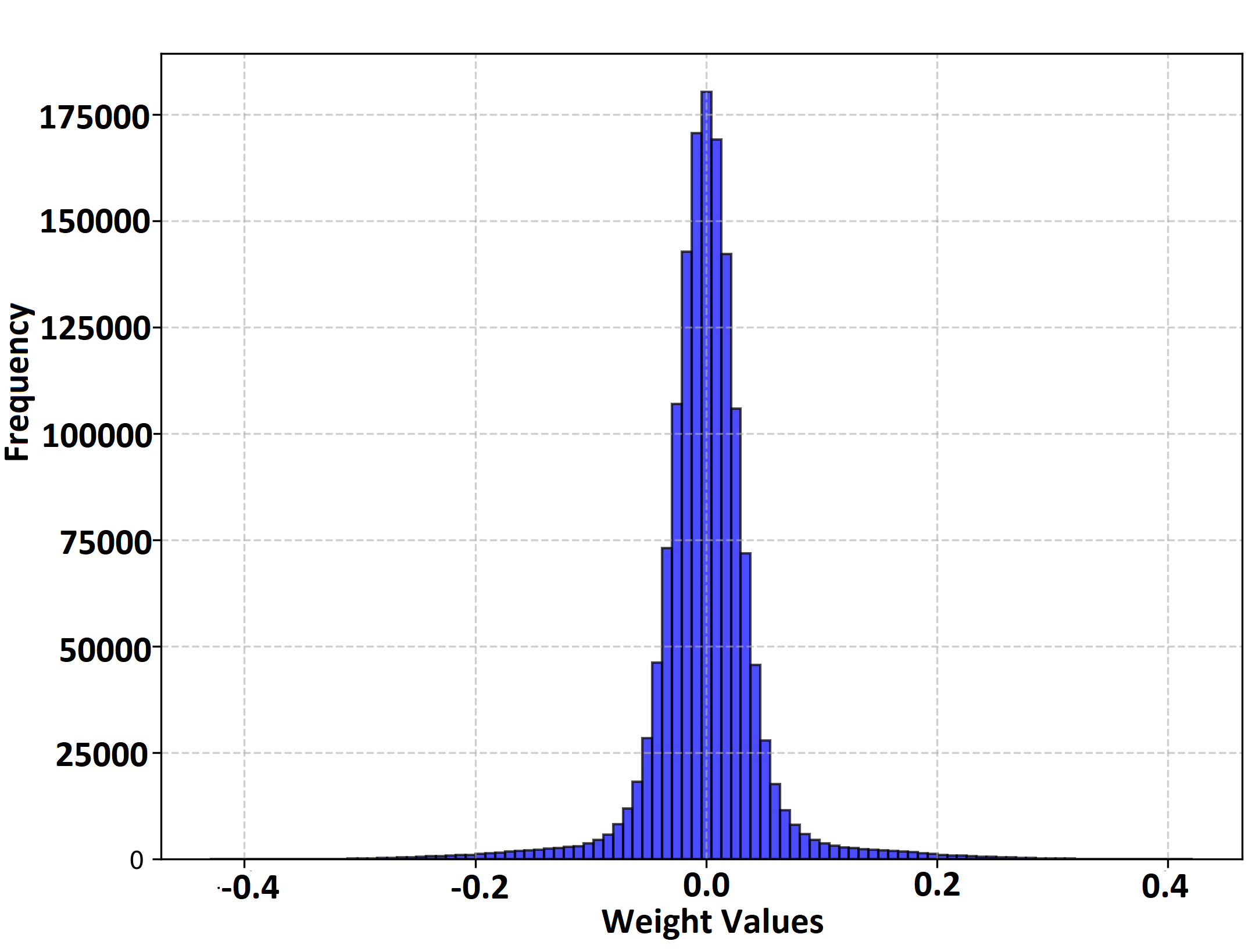}}
    \subfloat[Faulty model]{\includegraphics[width=0.32\textwidth]{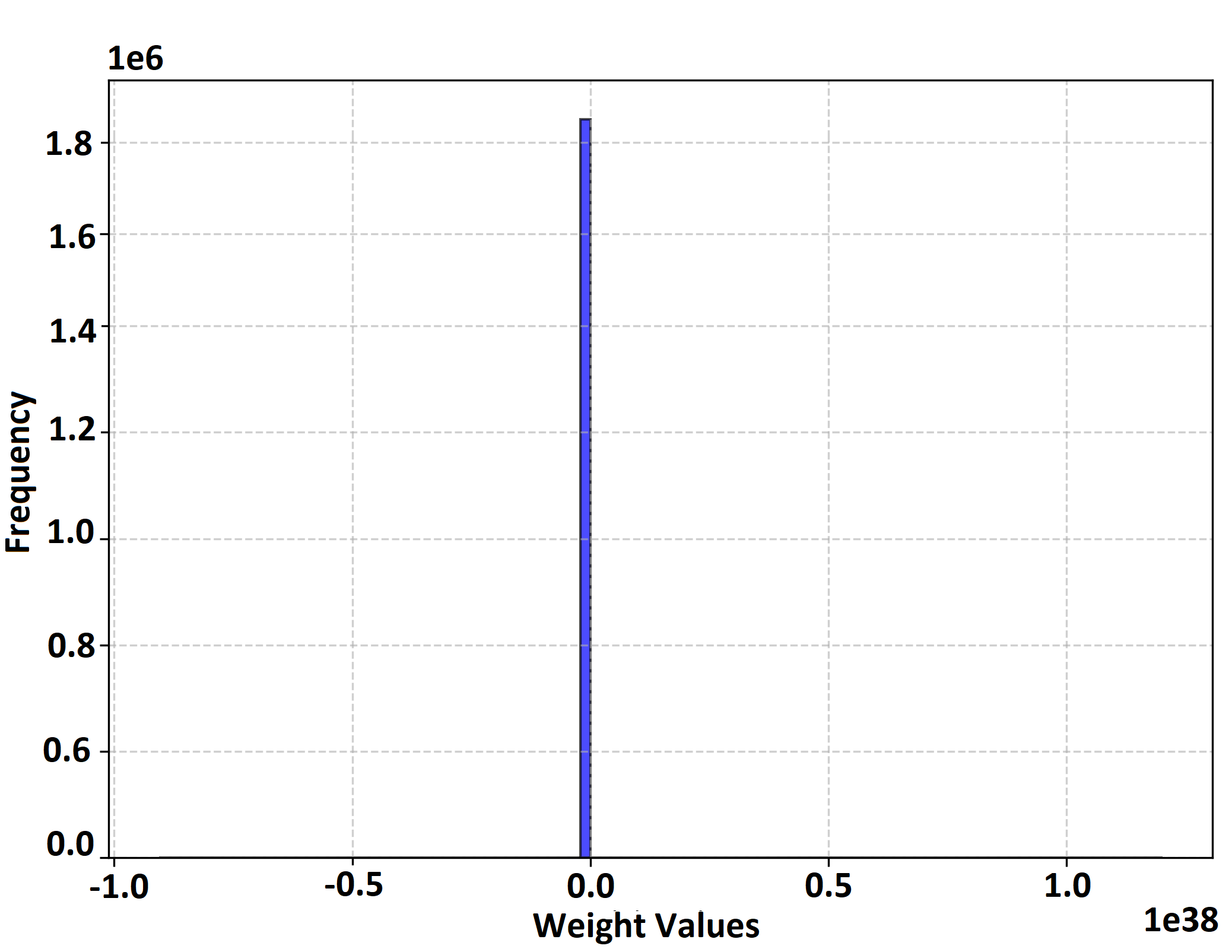}}
    \subfloat[Average]{\includegraphics[width=0.32\textwidth]{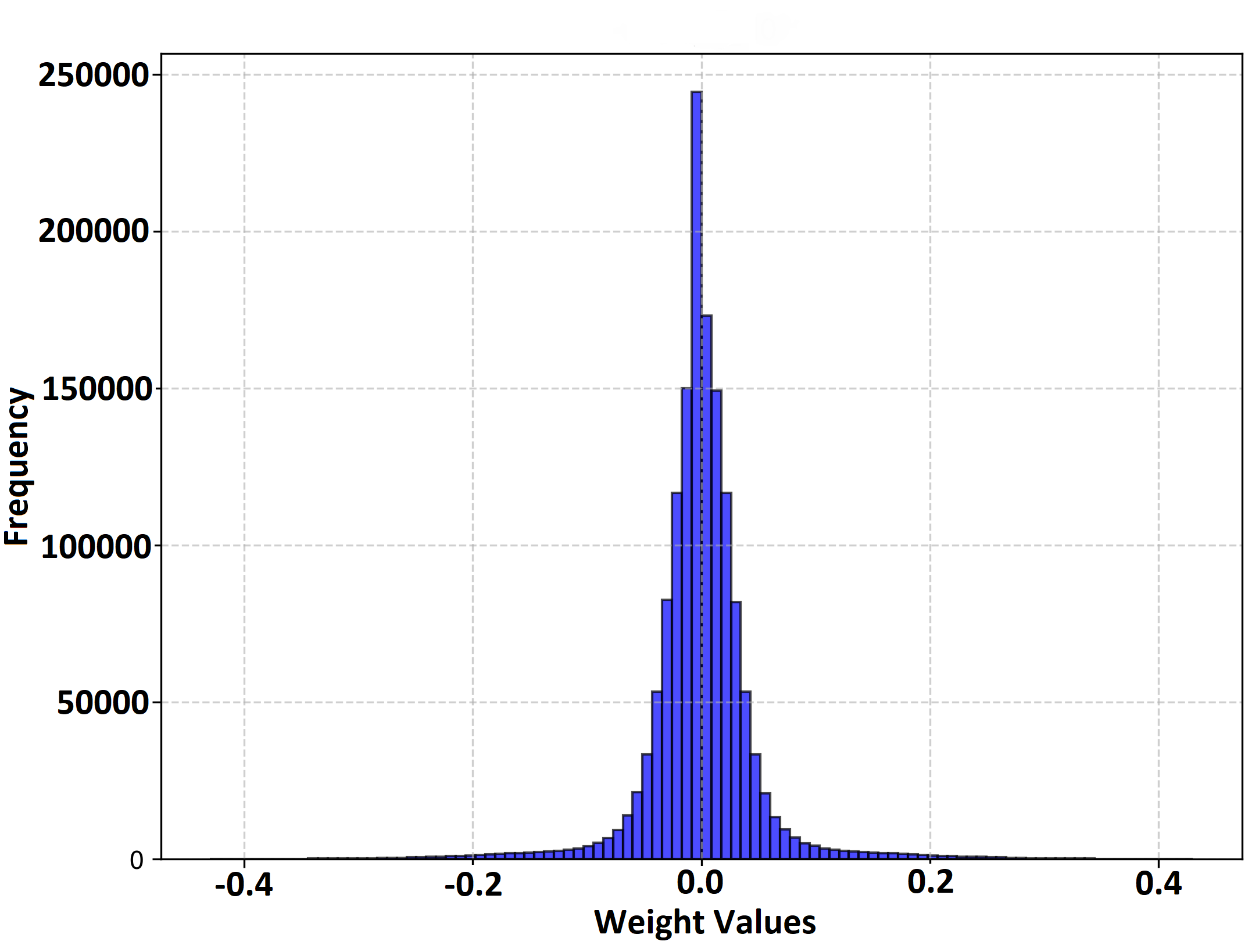}} \\
    \subfloat[MinMax]{\includegraphics[width=0.32\textwidth]{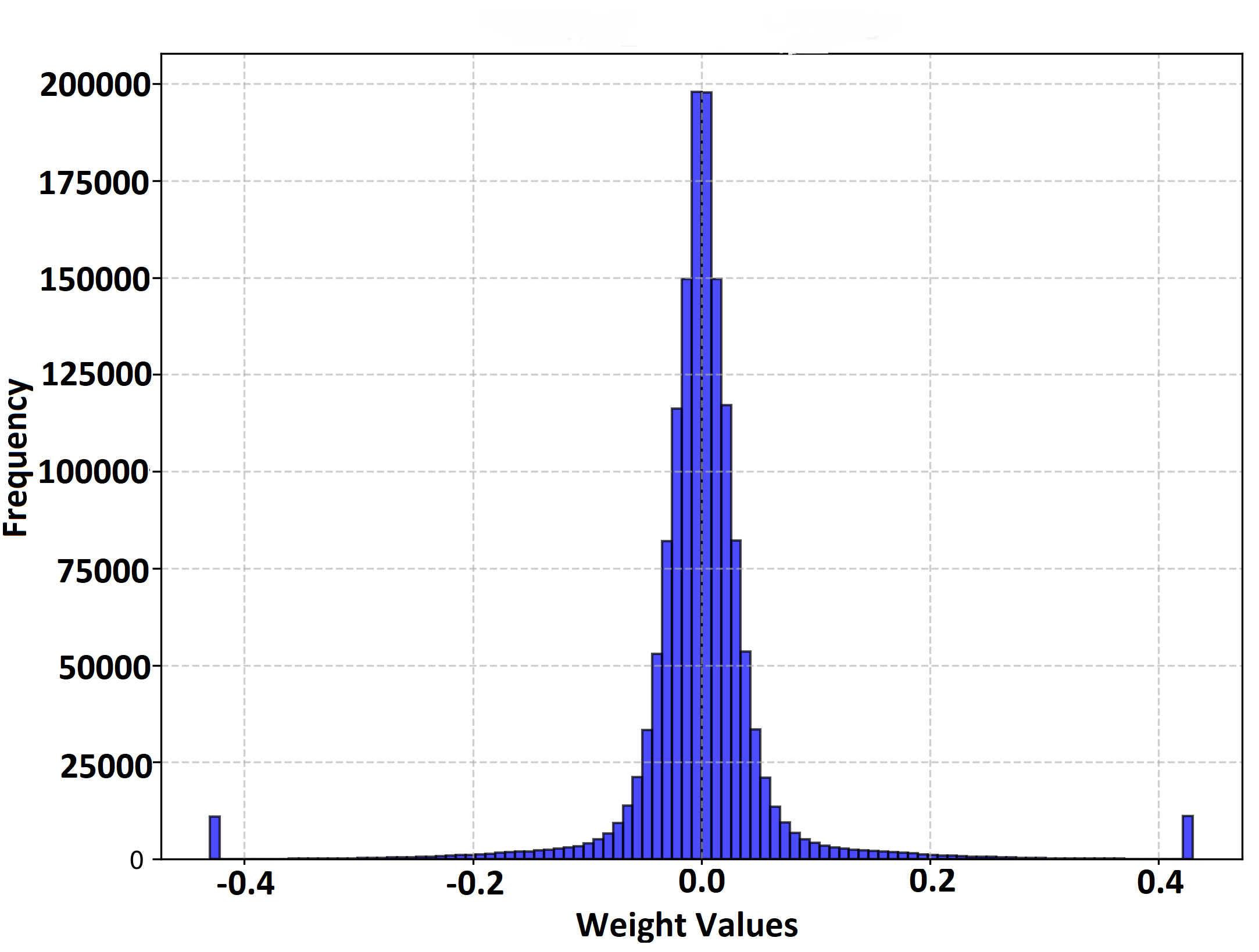}}
    \subfloat[WBC]{\includegraphics[width=0.32\textwidth]{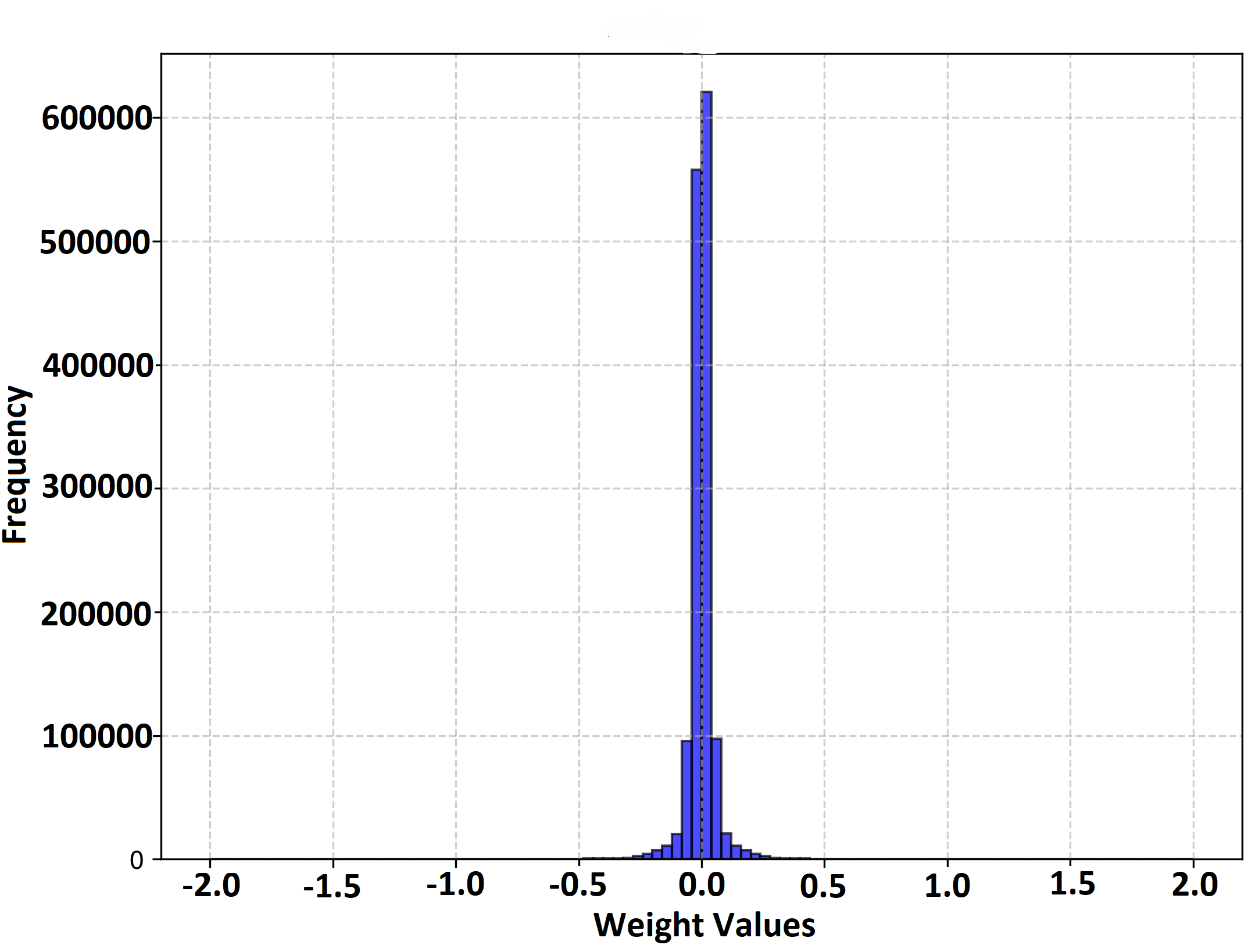}}
    \subfloat[CoG]{\includegraphics[width=0.32\textwidth]{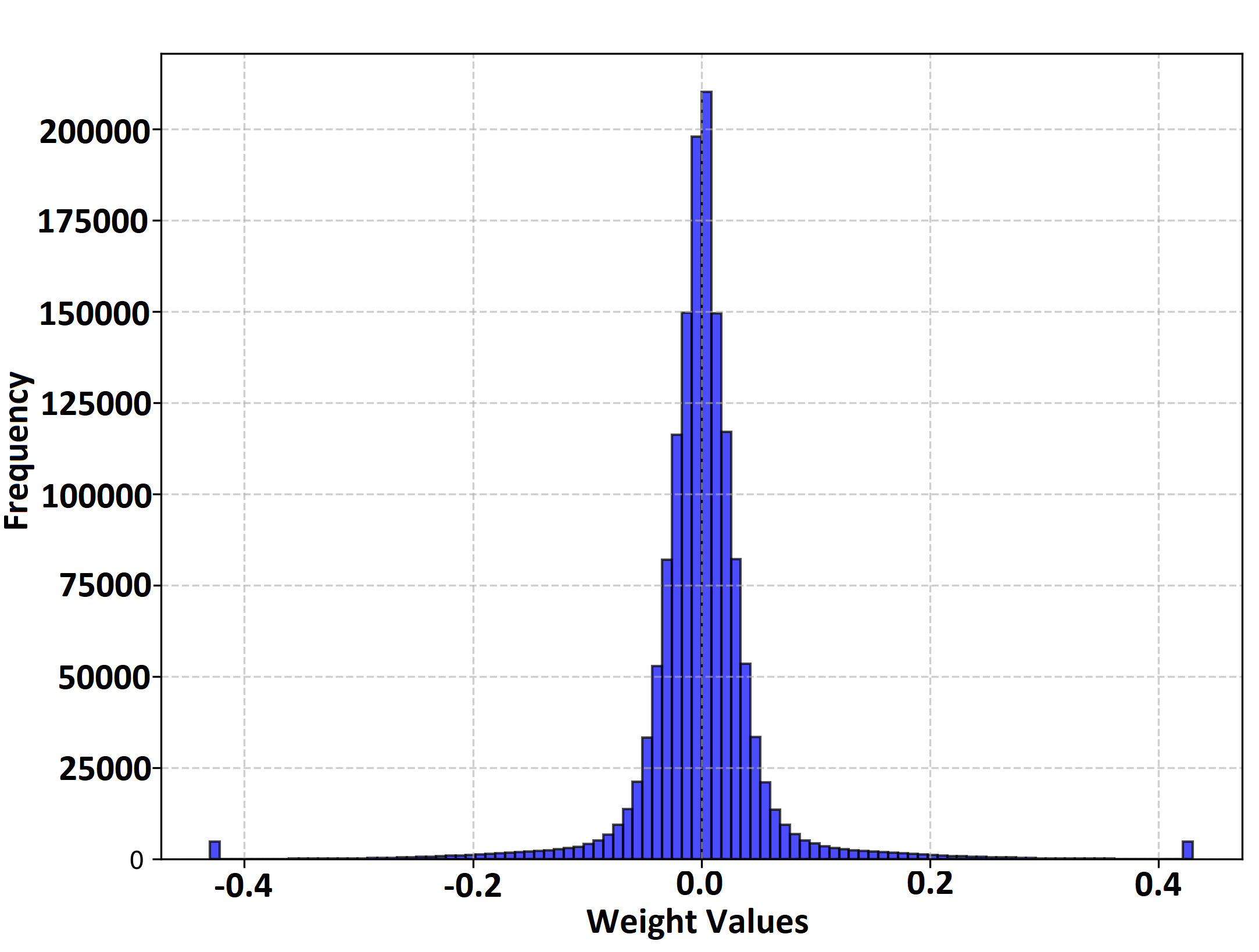}}
    \caption{Statistical distribution of weights in a representative StageNet layer under fault injection ($\text{BER}=0.01$). 
    (a) Golden baseline without fault, 
    (b) faulty model without protection, 
    (c) Average correction, 
    (d) MinMax clamping, 
    (e) Bit30 simplification, 
    and (f) the proposed CoG-based correction.}
    \label{fig:histogram}
\end{figure*}

Fig.~\ref{fig:histogram} illustrates the weight distribution of a representative StageNet layer under various protection schemes at $\text{BER}=0.01$.  
The fault-free distribution follows a near-normal shape, while the faulty model exhibits significant distortion. Each protection method attempts to restore the distribution toward its normal form; however, the proposed CoG-based correction most closely aligns with the original fault-free distribution. By effectively controlling both the central peak and boundary spikes, CoG demonstrates its ability to correct fault-induced deviations more precisely than statistical approaches. It is noted that AVC is not included in this comparison, as it operates on activation values rather than directly modifying weight distributions.

\subsection{Ablation Study on Convolutional Networks}
Applying the full CoG-based framework to deep convolutional networks introduces a significant computational overhead due to the large number of parameters and the high dimensionality of convolutional weight tensors. In preliminary experiments, extending the exhaustive distance optimization to VGG16 and ResNet18 resulted in execution times that exceeded the practical limits of the available hardware. This computational bottleneck makes the exhaustive variant unsuitable for large-scale CNNs under realistic deployment constraints.

\begin{table*}[htpb]
\centering
\caption{Comparison of AAA drop under different weight reliability protections for CNN models on the CIFAR-10 dataset. Lower values indicate better reliability against fault injection.}
\label{tab:dropaaa-cnn}

\begin{subtable}{0.48\textwidth}
\centering
\caption{VGG16}
\begin{tabular}{lrrrrrr}
\hline
Protection & 1E-08 & 1E-07 & 1E-06 & 1E-05 & 1E-04 & 1E-03 \\
\hline
$CoG_b$ & 0.00 & 0.00 & 0.00 & 0.00 & 0.07 & 0.76 \\
Average \cite{ahmadilivani2023analysis} & 0.06 & 0.06 & 0.06 & 0.08 & 0.14 & 0.80 \\
MinMax \cite{chen2021low}  & 0.02 & 0.02 & 0.02 & 0.03 & 0.12 & 2.17 \\
WBC \cite{parchekani2024zero}     & 0.02 & 0.02 & 0.02 & 0.04 & 0.39 & 6.12 \\
AVC \cite{parchekani2024zero}     & 0.41 & 0.97 & 7.25 & 47.29 & 77.24 & 92.50 \\
NPrt    & 1.01 & 6.69 & 47.24 & 92.48 & 92.50 & 92.50 \\
\hline
\end{tabular}
\end{subtable}
\hfill
\begin{subtable}{0.48\textwidth}
\centering
\caption{ResNet18}
\begin{tabular}{lrrrrrr}
\hline
Protection & 1E-08 & 1E-07 & 1E-06 & 1E-05 & 1E-04 & 1E-03 \\
\hline
$CoG_b$ & 0.10 & 0.10 & 0.10 & 0.11 & 0.13 & 0.17 \\
Average \cite{ahmadilivani2023analysis} & 0.11 & 0.11 & 0.11 & 0.11 & 0.13 & 0.25 \\
MinMax \cite{chen2021low}  & 0.14 & 0.14 & 0.14 & 0.14 & 0.17 & 0.48 \\
WBC \cite{parchekani2024zero}     & 0.14 & 0.14 & 0.14 & 0.26 & 0.73 & 4.18 \\
AVC \cite{parchekani2024zero}     & 7.51 & 55.21 & 92.32 & 92.32 & 92.32 & 92.32 \\
NPrt    & 9.71 & 56.51 & 92.32 & 92.32 & 92.32 & 92.32 \\
\hline
\end{tabular}
\end{subtable}

\end{table*}

To overcome this limitation, we employ the binary $CoG_b$ strategy, which eliminates exhaustive distance search while retaining the core principle of spatially informed weight correction. This design choice enables efficient evaluation on convolutional architectures and allows us to study the robustness of CoG-guided protection under constrained computational resources. The consistent reliability gains observed across all BERs indicate that the effectiveness of the CoG approach does not rely solely on exhaustive optimization, but rather on its ability to exploit spatial structure within weight tensors. As shown in Table~\ref{tab:dropaaa-cnn}, $CoG_b$ maintains strong reliability across all BERs.

%% file: sections/6-Conclusion.tex
\section{Conclusions}
\label{sec:conclusion}

This paper introduces a novel fault tolerance mechanism for DNNs deployed in safety-critical applications. The proposed method leverages the spatial patterns of weight matrices through the CoG concept to correct faulty parameters, ensuring both high reliability and minimal accuracy degradation under hardware faults.

A lightweight detection mechanism identifies faulty weights based on each layer's statistical distribution, followed by a spatial correction strategy that adjusts weights based on their proximity to the CoG. Validation on hybrid networks (StageNet and MTFNet) and conventional CNNs (ResNet-18, VGG-16) shows that the CoG-based approach outperforms traditional fault mitigation methods, providing superior resilience without significant computational overhead.

Experimental results demonstrate that CoG-based correction effectively preserves model accuracy under high fault rates, reduces error propagation, and maintains a weight distribution closer to the fault-free state compared to other methods. Specifically, under a BER of $10^{-3}$, the proposed method achieves fault tolerance improvements of up to $230\times$ and $6.41\times$ for StageNet and MTFNet, respectively, while providing up to $49.55\times$ and $20.79\times$ improvements on ResNet-18 and VGG-16 under comparable fault conditions. This fault tolerance approach is scalable, efficient, and applicable to a wide range of DNN architectures in safety-critical systems.